\documentclass[runningheads]{llncs}

\usepackage{graphicx}
\usepackage[backend=biber, style=numeric, maxbibnames=99, sorting=none]{biblatex}
\usepackage{multirow}
\usepackage{amssymb}
\usepackage{amsmath}
\usepackage{dsfont}
\usepackage[utf8]{inputenc}
\usepackage[T1]{fontenc}
\usepackage[english]{babel}
\usepackage{mathtools}
\usepackage[ruled]{algorithm2e}
\usepackage{algpseudocode}

\usepackage{subfig}

\usepackage[colorinlistoftodos,prependcaption]{todonotes}
\usepackage{blindtext}
\usepackage{hyperref}

\begin{document}

\title{Learning image quality assessment by reinforcing task amenable data selection}
\titlerunning{Image quality assessment by reinforcement learning}

\author{Shaheer U. Saeed\inst{1} \and
Yunguan Fu\inst{1, 2} \and
Zachary M. C. Baum\inst{1} \and
Qianye Yang\inst{1} \and
Mirabela Rusu\inst{3} \and
Richard E. Fan\inst{4} \and
Geoffrey A. Sonn\inst{3, 4} \and
Dean C. Barratt\inst{1} \and
Yipeng Hu\inst{1}
}
\authorrunning{S.U. Saeed et al.}

\institute{Centre for Medical Image Computing, Wellcome/EPSRC Centre for Interventional \& Surgical Sciences, and Department of Medical Physics \& Biomedical Engineering, University College London, London, UK \and
InstaDeep, London, UK \and
Department of Radiology, Stanford School of Medicine, Stanford, California, USA \and
Department of Urology, Stanford School of Medicine, Stanford, California, USA \\
\email{shaheer.saeed.17@ucl.ac.uk}
}

\maketitle            

\begin{abstract}
In this paper, we consider a type of image quality assessment as a task-specific measurement, which can be used to select images that are more amenable to a given target task, such as image classification or segmentation. We propose to train simultaneously two neural networks for image selection and a target task using reinforcement learning. A controller network learns an image selection policy by maximising an accumulated reward based on the target task performance on the controller-selected validation set, whilst the target task predictor is optimised using the training set. The trained controller is therefore able to reject those images that lead to poor accuracy in the target task. In this work, we show that the controller-predicted image quality can be significantly different from the task-specific image quality labels that are manually defined by humans. Furthermore, we demonstrate that it is possible to learn effective image quality assessment without using a ``clean'' validation set, thereby avoiding the requirement for human labelling of images with respect to their amenability for the task. Using $6712$, labelled and segmented, clinical ultrasound images from $259$ patients, experimental results on holdout data show that the proposed image quality assessment achieved a mean classification accuracy of $0.94\pm0.01$ and a mean segmentation Dice of $0.89\pm0.02$, by discarding $5\%$ and $15\%$ of the acquired images, respectively. The significantly improved performance was observed for both tested tasks, compared with the respective $0.90\pm0.01$ and $0.82\pm0.02$ from networks without considering task amenability. This enables image quality feedback during real-time ultrasound acquisition among many other medical imaging applications.


\end{abstract}

\section{Introduction}\label{sec:intro}
Image quality assessment (IQA) has been developed in the field of medical image computing and image-guided intervention as it is important to ensure that the intended diagnostic, therapeutic or navigational tasks can be performed reliably. It is intuitive that low-quality images can result in inaccurate diagnoses or measurements obtained from medical images \cite{qa_retina, qa_fetal}, but there has been little evidence that such corroboration can be quantified, between completion of a specific clinical application and a single general-purpose IQA methodology. Chow and Paramesran~\cite{qa_review} also pointed out that measures of image quality may not indicate diagnostic accuracy. We further argue that a general-purpose approach for medical image quality assessment is both challenging and potentially counter-productive. For example, various artefacts, such as reflections and shadows, may not be present near regions of clinical interest, yet a ``good quality'' image might still have inadequate field-of-view for the clinical task. In this work, we investigate the type of image quality which indicates how well a specific downstream target task performs and refer to this quality as \textit{task amenability}. 

Current IQA approaches in clinical practice rely on subjective human interpretation of a set of \textit{ad hoc} criteria~\cite{qa_review}. Automating IQA methods, for example, by computing dissimilarity to empirical references~\cite{qa_review}, typically can provide an objective and repeatable measurement, but requires robust mathematical models to approximate the underlying statistical and physical principles of good-quality image generation process or known mechanisms that reduce image quality (e.g. \cite{us_qa_carotid, qa_vessel}). Recent deep-learning-based IQA approaches provide fast inference using expert labels of image quality for training \cite{qa_fetal, qa_retina_deep, qa_liver_deep, qa_zachary_lung}. However, besides the potentially high variability in these human-defined labels, to what extent they reflect task amenability - i.e. their usefulness for a specific task - is still an open question. In particular, a growing number of these target tasks have been modelled and automated by, for example, neural networks, which may result in different or unknown task amenability.

In this work, we focus on a specific use scenario of the task-specific IQA, in which images are selected by the measured task-specific image quality, such that the selected subset of high-quality images leads to improved target classification or segmentation accuracy. This image selection by task amenability has many clinical applications, such as meeting a clinically-defined accuracy requirement by removing the images with poor task amenability and maximising task performance given a predefined tolerance on how many images with poor amenability can be rejected and discarded. The rejected images may be re-acquired immediately in applications such as the real-time ultrasound imaging investigated in this work. The IQA feedback during scanning also provides an indirect measure of user skills, though skill assessment is not discussed further in this paper.

Furthermore, we propose to train a \textit{controller network} and a \textit{task predictor network} together for selecting task amenable images and for completing the target task, respectively. We highlight that optimising the controller is dependent on the task predictor being optimised. This may therefore be considered a meta-learning problem that maximises the target task performance with respect to the controller-selected images. 

Reinforcement learning (RL) has increasingly been used for meta-learning problems, such as augmentation policy search \cite{autoaugment, adv_autoaugment}, automated loss function search \cite{autoloss} and training data valuation \cite{google_dvrl}. Common in these approaches, a target task is optimised with a controller which modifies parameters associated with this target task. The parameter modification action is followed by a reward signal computed based on the target task performance, which is subsequently used to optimise the controller. This allows the controller to learn the parameter setting that results in a better performed target task. The target application can be image classification, regression or segmentation, while the task-associated parameter modification actions include transforming training data for data augmentation \cite{autoaugment, adv_autoaugment}, selecting convolution filters and activation functions for network architecture search \cite{archisearch} and sampling training data for data valuation \cite{google_dvrl}. Among these recent developments, the data valuation approach \cite{google_dvrl} shares some interesting similarities with our proposed IQA method, but with several important differences in the reward formulation by weighting/sampling validation set, the availability of ``clean'' high-quality image data, in addition to the different RL algorithms and other methodological details described in Sec. \ref{sec:method}. For medical imaging applications, the RL-based meta-learning has also been proposed, for instance, to search for optimal weighting between different ultrasound modalities for the downstream breast cancer detection \cite{autoweight} and to optimise hyper-parameters for a subsequent 3D medical image segmentation~\cite{yang2019searching}, using the REINFORCE algorithm \cite{REINFORCE} and the proximal policy optimization algorithm~\cite{schulman2017proximal}, respectively.

In this work, we propose using RL to train the controller and the task predictor for assessing medical image quality with respect to two common medical image analysis tasks. Using medical ultrasound data acquired from prostate cancer patients, the two tasks are a) classifying 2D ultrasound images that contain prostate glands from those that do not and b) segmenting the prostate gland. These two tasks are not only the basis of several computational applications, such as 3D volume reconstruction, image registration and tumour detection, but are also directly useful for navigating ultrasound image acquisition during surgical procedures, such as ultrasound-guided biopsy and therapies. Our experiments were designed to investigate the following research questions:

\begin{itemize}
\item Can the task performance be improved on holdout test data selected by the trained controller network, compared with the same task predictor network based on supervised training and non-selective test data?
\item Does the trained controller network provide a better or different measure of task amenability, compared with human labels of image quality that are intended to indicate amenability to the same tasks?
\item What is the trade-off between the quantity of rejected images and the improvement in task performance? 
\end{itemize}

The contributions are summarised as follows: We 1) propose to formulate task-specific IQA to learn task amenable data selection; 2) propose a novel RL-based approach to quantify the task amenability, using different reward formulations with and without the need for human labels of task amenability; and 3) present experiments to demonstrate the efficacy of the proposed IQA approach using real medical ultrasound images in two different downstream target tasks.

\begin{figure}[!ht]
\centering
\includegraphics[width=\textwidth]{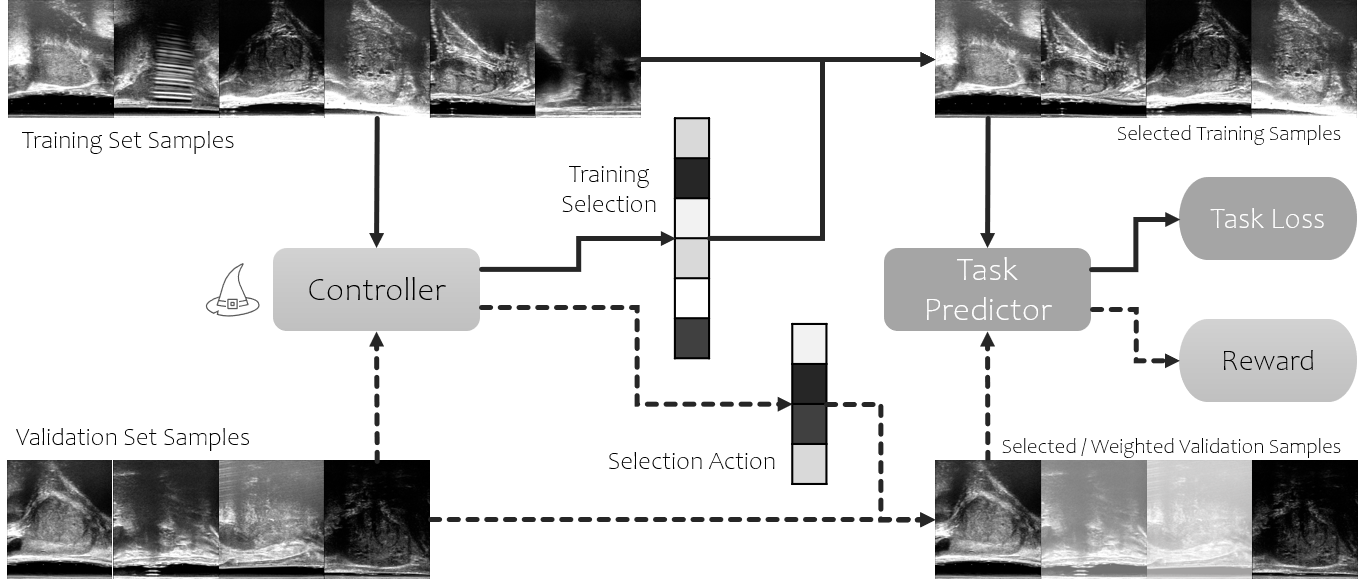}
\caption{Illustration of the training for controller and task predictor networks.}
\label{fig:method}
\end{figure}

\section{Method}\label{sec:method}

\subsection{Image quality assessment by task amenability}

The proposed IQA consists of two parametric functions, task predictor and controller, illustrated in Fig. \ref{fig:method}. The task predictor $f(\cdot;w):\mathcal{X}\rightarrow\mathcal{Y}$, with parameters $w$, outputs a prediction $y\in\mathcal{Y}$ for a given image sample $x\in\mathcal{X}$. The controller $h(\cdot;\theta):\mathcal{X}\rightarrow[0,1]$, with parameters $\theta$, generates an image quality score for a sample $x$, measuring task amenability of the sample. $\mathcal{X}$ and $\mathcal{Y}$ denote the image and label domains specific to a certain task, respectively. 

Let $\mathcal{P}_{X}$ and $\mathcal{P}_{XY}$ be the image distribution and the joint image-label distribution, with probability density functions $p(x)$ and $p(x,y)$, respectively. The task predictor's objective is to minimise a weighted loss function $L_f:\mathcal{Y}\times\mathcal{Y}\rightarrow\mathbb{R}_{\geq0}$:
\begin{align}
    \min_w \mathbb{E}_{(x,y)\sim\mathcal{P}_{XY}}[L_f(f(x;w), y)h(x;\theta)],
\end{align}
where $L_f$ measures how well the task is performed by the predictor $f(x;w)$, given label $y$. It is weighted by the controller-measured task amenability on the same image $x$, as mistakes (high loss) on images with lower task amenability ought to be less weighted - with a view to rejecting them, and vice versa. The controller's objective is to minimise a weighted metric function $L_h:\mathcal{Y}\times\mathcal{Y}\rightarrow\mathbb{R}_{\geq0}$:
\begin{align}
    \min_\theta \mathbb{E}_{(x,y)\sim\mathcal{P}_{XY}}[L_h(f(x;w), y)h(x;\theta)],\\
    \text{s.t.}\quad \mathbb{E}_{x\sim\mathcal{P}_{X}}[h(x;\theta)] \geq c > 0
\end{align}
such that the controller is encouraged to predict lower quality scores for images with higher metric values (lower task performance), as the weighted sum is minimised. The intuition is that making correct predictions on low-quality images tends to be more difficult. The constraint prevents the trivial solution $h\equiv0$.

Thus, the overall objective to learn the proposed task-specific IQA can be assembled as the following minimisation problem:
\begin{subequations}\label{eq:iqa-weighted}
\begin{align}
    && \min_\theta \mathbb{E}_{(x,y)\sim\mathcal{P}_{XY}}[L_h(f(x;w^*), y)h(x;\theta)],\\
    \text{s.t.}&& w^*=\arg\min_w \mathbb{E}_{(x,y)\sim\mathcal{P}_{XY}}[L_f(f(x;w), y)h(x;\theta)],\\
    &&\mathbb{E}_{x\sim\mathcal{P}_{X}}[h(x;\theta)] \geq c > 0.
\end{align}
\end{subequations}
To facilitate a sampling or selection action (see Sec. \ref{subsec:iqa-rl}) by controller-predicted task amenability scores, Eq.~\eqref{eq:iqa-weighted} is re-written as:

\begin{subequations}\label{eq:iqa-sampled}
\begin{align}
    && \min_\theta \mathbb{E}_{(x,y)\sim\mathcal{P}_{XY}^h}[L_h(f(x;w^*), y)],\\
    \text{s.t.}&& w^*=\arg\min_w \mathbb{E}_{(x,y)\sim\mathcal{P}_{XY}^h}[L_f(f(x;w), y)],\\
    &&\mathbb{E}_{x\sim\mathcal{P}_{X}^h}[1] \geq c > 0.
\end{align}
\end{subequations}
where the data $x$ and $(x,y)$ are sampled from the controller-selected or -sampled distributions $\mathcal{P}_{X}^h$ and $\mathcal{P}_{XY}^h$, with probability density functions $p^h(x) \propto p(x)h(x;\theta)$ and $p^h(x,y) \propto p(x,y)h(x;\theta)$, respectively.

\subsection{The reinforcement learning algorithm}\label{subsec:rl}

In this work, an RL agent interacting with an environment is considered as a finite-horizon Markov decision process with $(\mathcal{S}, \mathcal{A}, p, r, \pi, \gamma)$. $\mathcal{S}$ is the state space and $\mathcal{A}$ is a continuous action space. $p:\mathcal{S}\times\mathcal{S}\times\mathcal{A}\rightarrow[0,1]$ is the state transition distribution conditioned on state-actions, e.g. $p(s_{t+1}\mid s_t,a_t)$ denotes the probability of the next state $s_{t+1}\in\mathcal{S}$ given the current state $s_t\in\mathcal{S}$ and action $a_t\in\mathcal{A}$. $r:\mathcal{S}\times\mathcal{A}\rightarrow\mathbb{R}$ is the reward function and $R_t=r(s_t,a_t)$ denotes the reward given $s_t$ and $a_t$. $\pi(a_t\mid s_t):\mathcal{S}\times\mathcal{A}\in[0,1]$ is the policy represents the probability of performing action $a_t$ given $s_t$. The constant $\gamma\in[0,1]$ discounts the accumulated rewards starting from time step $t$: $Q^\pi(s_t, a_t) = \sum_{k=0}^{T}\gamma^kR_{t+k}.$
A sequence $(s_1, a_1, R_1, s_2, a_2, R_2, \ldots,  s_T, a_T, R_T)$ is thereby created with the RL agent training, with the objective to learn a parameterised policy $\pi_\theta$ which maximises the expected return $J(\theta)=\mathbb{E}_{\pi_\theta}[Q^\pi(s_t, a_t)].$

Two different algorithms have been considered in our experiments, REINFORCE \cite{REINFORCE} and Deep Deterministic Policy Gradient (DDPG) \cite{ddpg}. Based on initial results indicating little difference in performance between the two, all the results presented in this paper are based on DDPG, with which a noticeably more efficient and stable training was observed. Further investigation into the choice of RL algorithms remains interesting in future work. While the REINFORCE computes policy gradient to update the controller parameters directly, DDPG is an actor-critic algorithm, with an off-policy critic $Q(s_t,a_t;\theta^Q):\mathcal{S}\times\mathcal{A}\rightarrow\mathbb{R}$ and a deterministic actor $\mu(s_t;\theta^\mu):\mathcal{S}\rightarrow\mathcal{A}$. To maximise the performance function $J(\theta^\mu)=\mathbb{E}_{\mu}[Q^\pi(s_t, \mu(s_t;\theta^\mu))]$, the variance-reduced policy gradient is used to update the controller:
$\nabla_{\theta^\mu} J(\theta^\mu)=\mathbb{E}_{\mu}[\nabla_{\theta^\mu}\mu(s_t;\theta^\mu)\nabla_aQ^\pi(s_t, a)|_{a=\mu(s_t;\theta^\mu)}]$, which can be approximated by sampling the behaviour policy $\beta(s_t)\neq\mu(s_t;\theta^\mu)$:
\begin{align}
 \nabla_{\theta^\mu} J(\theta^\mu)\approx\mathbb{E}_{\beta}[\nabla_{\theta^\mu}\mu(s_t;\theta^\mu)\nabla_aQ(s_t, a;\theta^Q)|_{a=\mu(s_t;\theta^\mu)}],
\end{align}
where the critic $Q(s_t, a_t;\theta^Q)$ is updated with respect to minimising:
\begin{align}
 \mathbb{E}_\beta[(R_t+\gamma Q(s_{t+1}, a_{t+1};\theta^Q) - Q(s_t, a_t;\theta^Q))^2].
\end{align}
In our implementation, copies of the actor $Q'(s_t, a_t; \theta^{Q'})$ and the critic $\mu'(s_t; \theta^{\mu'})$ are used for computing moving averages during parameter updates, $\theta^{Q'} \leftarrow \tau \theta^{Q} + (1-\tau) \theta^{Q'}$ and $\theta^{\mu'} \leftarrow \tau \theta^{\mu} + (1-\tau) \theta^{\mu'}$, respectively. Additionally, a random noise $\mathcal{N}$ is added to $\mu(s_t;\theta^{\mu})$ for exploration. Here, $\tau=0.001$ and $\mathcal{N}$ is the Ornstein-Uhlenbeck process \cite{ornstein_noise} with the scale and the mean reversion rate parameters set to 0.2 and 0.15, respectively.

\subsection{Image quality assessment with reinforcement learning}\label{subsec:iqa-rl}

In this section, the IQA in Eq.\eqref{eq:iqa-sampled} is formulated as a RL problem and solved by the algorithm described in Sec. \ref{subsec:rl}. The pseudo-code is provided in Algorithm \ref{algo:iqa}. A finite dataset together with the task predictor is considered the environment. At time step $t$, the observed state from the environment $s_t=(f(\cdot;w_t), \mathcal{B}_t)$ consists of the predictor $f(\cdot;w_t)$ and a mini-batch of samples $\mathcal{B}_t=\{(x_i,y_i)\}_{i=1}^B$ from a training dataset $\mathcal{D}_\text{train}=\{(x_i,y_i)\}_{i=1}^N$. The agent is the controller $h(\cdot;\theta)$ that outputs sampling probabilities $\{h(x_i;\theta)\}_{i=1}^B$. The action $a_t=\{a_t^i\}_{i=1}^B\in\{0,1\}^B$ is the sample selection decision, by which $(x_i,y_i)$ is selected if $a_t^i=1$ for training the predictor. The policy $\pi_\theta(a_t\mid s_t)$ is thereby defined as:
\begin{align}
    \log\pi_\theta(a_t\mid s_t)=\sum_{i=1}^{D}h(x_i;\theta)a_t^i + (1-h(x_i;\theta)(1-a_t^i))
\end{align}
The unclipped reward $\tilde{R}_t$ is calculated based on the predictor's performance $\{l_{j,t}\}_{j=1}^M = \{L_h(f(x_j;w_t), y_j)\}_{j=1}^M$ on a validation dataset $\mathcal{D}_\text{val}=\{(x_j,y_j)\}_{j=1}^M$ and the controller's outputs $\{h_j\}_{j=1}^M = \{h(x_j;\theta)\}_{j=1}^M$. Three definitions for reward computation are considered in this work:
\begin{enumerate}
    \item $\tilde{R}_{\text{avg},t}=-\frac{1}{M}\sum_{j=1}^Ml_{j,t}$, the average performance.
    \item $\tilde{R}_{\text{w},t}=-\frac{1}{M}\sum_{j=1}^Ml_{j,t}h_j$, the weighted sum.
    \item $\tilde{R}_{\text{sel},t}=-\frac{1}{M'}\sum_{j'=1}^{M'}l_{j',t}$, the average of the selected $M'$ samples.
\end{enumerate}
where $\{{j'}\}_{j'=1}^{M'} \subseteq \{j\}_{j=1}^M$ and $h_{j'}\leq h_{k'}, \forall{k'\in\{j'\}^c},\forall{j'\in\{j'\}}$, i.e. the unclipped reward $\tilde{R}_{\text{sel},t}$ is the average of $\{l_{j'}\}$ from the subset of $M'=\lfloor (1-s^{rej})M \rfloor$ samples, by removing the first $s^{rej}\times100\%$ samples, after sorting $h_j$ in decreasing order. It is important to note that, for the first reward definition $\tilde{R}_{\text{avg},t}$ without being weighted or selected by the controller, the validation set requires pre-selected ``high-amenability'' data. In this work, additional human labels of task amenability were used for generating such a clean fixed validation set (details in Sec. \ref{sec:exp}). During training, the clipped reward $R_t=\tilde{R}_t-\bar{R}_t$ is used with a moving average $\bar{R}_t={\alpha}_R\bar{R}_{t-1}+(1-{\alpha}_R)\tilde{R}_t$, where ${\alpha}_R$ is a hyper-parameter set to 0.9.

\begin{algorithm}[!ht]
    \SetAlgoLined
    \KwData{Training dataset $\mathcal{D}_\text{train}$ and validation dataset $\mathcal{D}_\text{val}$.}
    \KwResult{Task predictor $f(\cdot;w)$ and controller $h(\cdot;\theta)$.}
    \BlankLine
    \While{not converged}{
        \For{$k\leftarrow 1$ \KwTo $K$}{
            \For{$t\leftarrow 1$ \KwTo $T$}{
                Sample a mini-batch $\mathcal{B}_t=\{(x_i,y_i)\}_{i=1}^{B}$ from $\mathcal{D}_\text{train}$\;
                Compute selection probabilities $\{h(x_i;\theta_t)\}_{i=1}^B$\;
                Sample actions $a_t=\{a_t^i\}_{i=1}^B$ w.r.t. $a_t^i\sim\text{Bernoulli}(h(x_i; \theta))$\;
                Selected samples $\mathcal{B}_{t,\text{selected}}$ from $\mathcal{B}_t$\;
                Update predictor $f(\cdot;w_t)$\ using $\mathcal{B}_{t,\text{selected}}$\;
                Compute reward $R_t$\;
            }
            Collect one episode $\{\mathcal{B}_t,a_t,R_t\}_{t=1}^T$\;
        }
        Update controller $h(\cdot;\theta)$ using reinforcement learning algorithm\;
    }
\caption{Image quality assessment by task amenability}\label{algo:iqa}
\end{algorithm}

\begin{figure}[!ht]
\centering
\includegraphics[width=\textwidth]{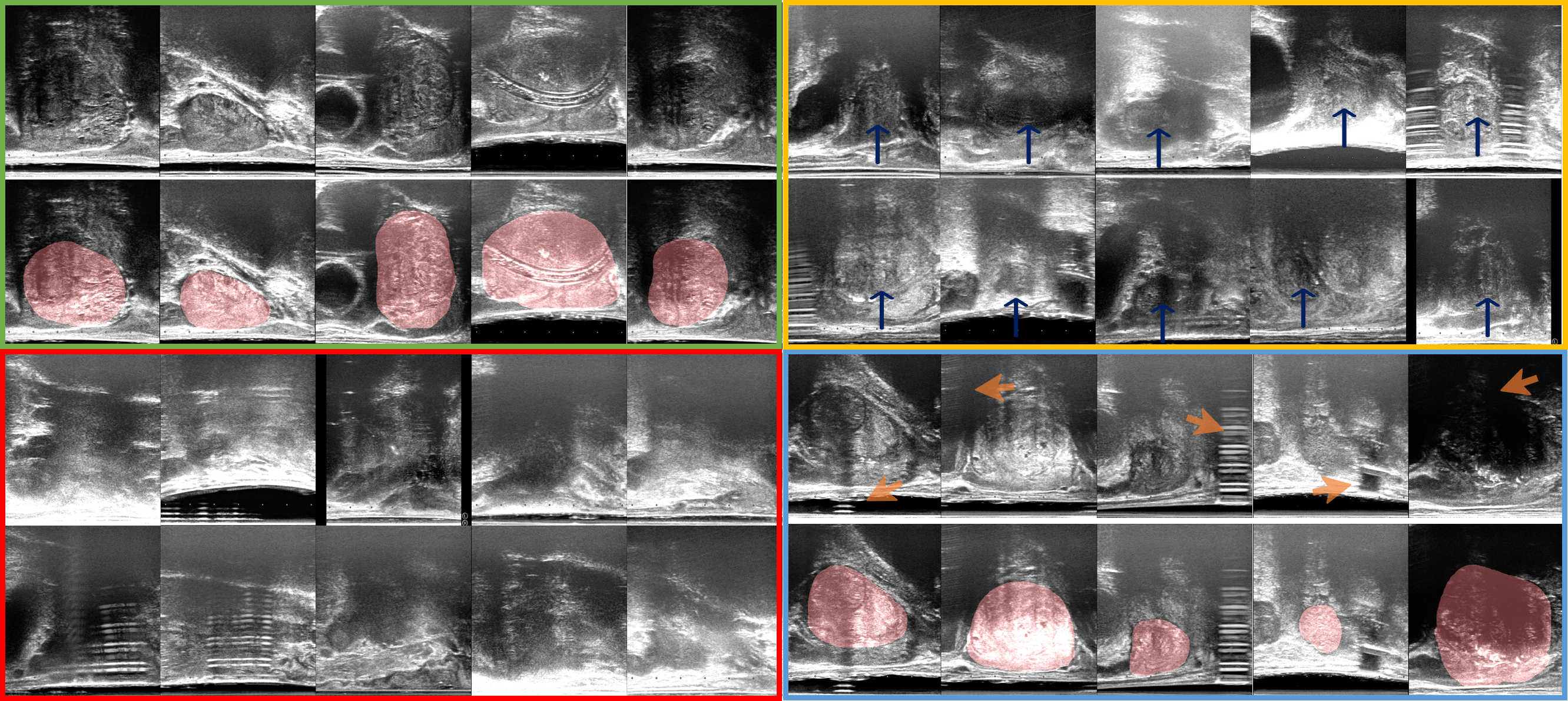}
\caption{Examples of ultrasound images in this study. \textbf{Top-left} (green): task-amenable images that contain prostate gland (shaded in red); \textbf{Bottom-left} (red): images with poor task amenability that are difficult to recognise prostate glands (for classification) and their boundaries (for segmentation); \textbf{Top-right} (yellow), images that are likely to contain prostate glands (blue arrows) but identifying the complete gland boundaries for segmentation is challenging; and \textbf{Bottom-right} (blue): images that contain visible noise and artefacts (orange arrows), but may be amenable to both classification and segmentation tasks.}
\label{fig:example_image}
\end{figure}

\section{Experiment}\label{sec:exp}

Transrectal ultrasound images were acquired from $259$ patients, at the beginning stages of the ultrasound-guided biopsy procedures, as part of the SmartTarget: THERAPY and SmartTarget: BIOPSY clinical trials (clinicaltrials.gov identifiers NCT02290561 and NCT02341677 respectively). For each subject, a range of 50-120 2D frames were acquired with the side-firing transducer of a bi-plane transperineal ultrasound probe (C41L47RP, HI-VISION Preirus, Hitachi Medical Systems Europe), during manual positioning a digital transperineal stepper (D\&K Technologies GmbH, Barum, Germany) or rotating the stepper with recorded relative angles, for navigating ultrasound view and scanning entire gland, respectively. For the purpose of feasibility in manual labelling, the ultrasound images were further sampled at approximately every $4$ degrees, resulting in $6712$ images in total.

Prostate glands were segmented in all images by three trained biomedical engineering researchers, in which the prostate gland is visible. Two sets of task labels were curated for individual images: classification labels (a binary scalar indicating the presence of prostate) and segmentation labels (a binary mask of the gland). In this work, a single label for each of the classification and segmentation tasks was obtained by consensus over all three observers, based on majority voting at image-level and pixel-level, respectively.

As discussed in Sec. \ref{sec:intro}, the task-specific image quality of interest for the classification task and the segmentation task can be different. Therefore, additional two binary labels were assigned for each image to represent the human label of task amenability, based on the observer assessment of whether the image quality adversely affects the completion of each task (see examples in Fig. \ref{fig:example_image}).

The labelled images were randomly split, at the patient-level, into train, validation, and holdout sets with $4689$, $1023$, and $1000$ images from $178$, $43$, and $38$ subjects, respectively.

The proposed RL framework was evaluated on both tasks. The three reward definitions proposed in Sec. \ref{subsec:iqa-rl} were compared together with two \textit{non-selective baseline} networks for classification and segmentation trained on all training data. For comparison purposes, they share the same network architectures and training strategies as the task predictors in the RL algorithms. For the classification tasks, Alex-Net \cite{alexnet_class,comp_dl} was trained with a cross-entropy loss and a reward based on classification accuracy (Acc.), i.e. classification correction rate. For segmentation tasks, U-Net \cite{unet_seg} was trained with a pixel-wise cross-entropy loss and a mean binary Dice score to form the reward. For the purpose of this work, the reported experimental results are based on empirically configured networks and RL hyper-parameters that were unchanged, unless specified, from the default values in the original Alex-Net, U-Net and DDPG algorithms. It is perhaps noteworthy that, based on our initial experiments, changing these configurations seems unlikely to alter the conclusions summarised in Sec. \ref{sec:result}, but future research may be required to confirm this and further optimise their performance.

Based on the holdout set, a mean Acc. and a mean binary Dice were computed to evaluate the trained task predictor networks, in classification and segmentation tasks, respectively, with different percentages of the holdout set removed according to the trained controller networks. Selection is not applicable to the baseline networks. Standard deviation (St.D.) is also reported to measure the inter-patient variance. Paired two-sample t-test results at a significance level of $\alpha=0.05$ are reported for comparisons.

\section{Result}\label{sec:result}

\begin{figure}[!ht]
\centering
\includegraphics[width=\textwidth]{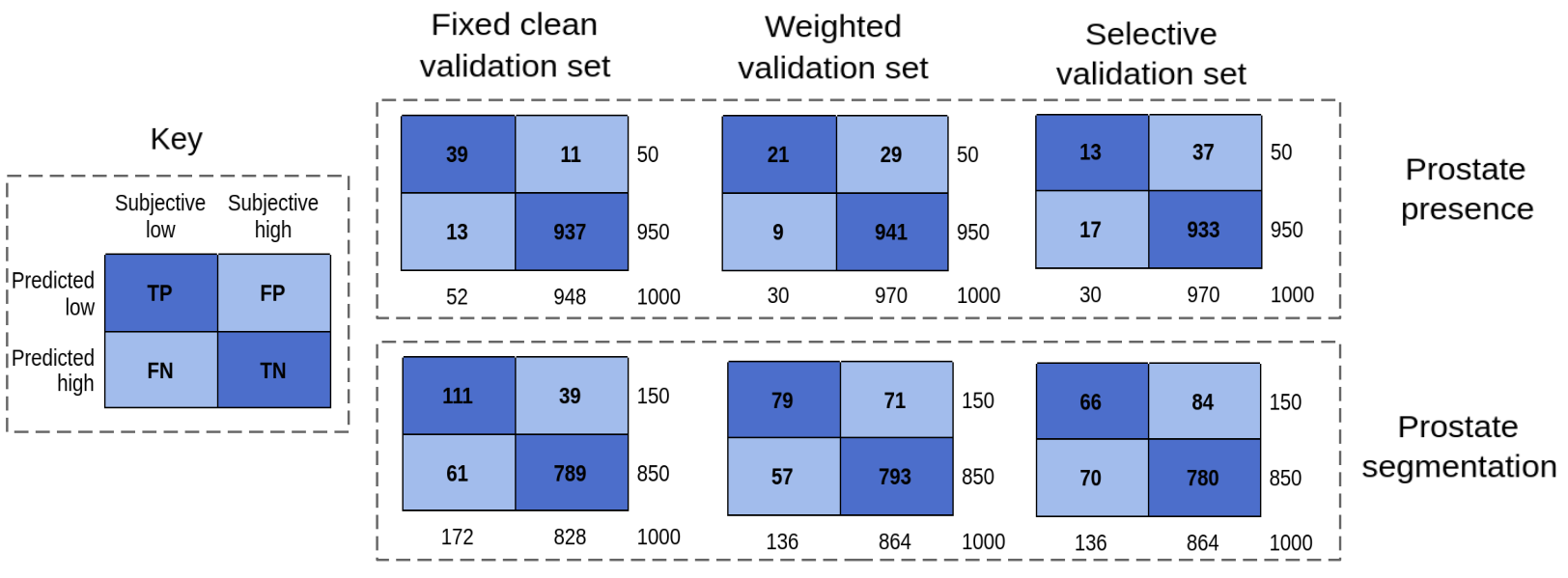}
\caption{Contingency tables comparing subjective labels to controller predictions for the different reward computation strategies.}
\label{fig:cm}
\end{figure}

To evaluate the trained controllers, the $2\times2$ contingency tables in Fig. \ref{fig:cm} compare subjective task amenability labels with controller predictions. For the purpose of comparison, $5\%$ and $15\%$ of images were removed from the holdout set by the trained controller, for the classification and segmentation tasks, respectively. The results of the selective reward $\tilde{R}_{\text{sel},t}$ with $s^{rej}=5\%$ and $s^{rej}=15\%$ are used as examples, for the two respective tasks. Thereby, agreement and disagreement are quantified between images assessed by the proposed IQA and the same images assessed by the subjective human labels of task amenability, denoted as predicted low/high and subjective low/high, respectively. In classifying prostate presence, the rewards based on fixed-, weighted- and selective validation sets resulted in agreed $75\%$, $70\%$ and $43\%$ low task amenability samples, with Cohen's kappa values of $0.75$, $0.51$ and $0.30$, respectively. In the segmentation task, the three rewards have $65\%$, $58\%$ and $49\%$ agreed low task amenability samples, with Cohen's kappa values of $0.63$, $0.48$ and $0.37$, respectively.

\begin{table}[!ht]
\centering
\begin{tabular}{|c|c|c|}
\hline
Task & Reward computation strategy & Mean $\pm$ St.D.\\
\hline
\multirow{4}{6em}{Prostate presence (Acc.)} & Non-selective baseline & 0.897 $\pm$ 0.010\\
\cline{2-3}
& $\tilde{R}_{\text{avg},t}$, fixed validation set & 0.935 $\pm$ 0.014\\
\cline{2-3}
& $\tilde{R}_{\text{w},t}$, weighted validation set & 0.926 $\pm$ 0.012\\
\cline{2-3}
& $\tilde{R}_{\text{sel},t}$, selective validation set & 0.913 $\pm$ 0.012\\
\hline
\multirow{4}{6em}{Prostate segmentation (Dice)} & Non-selective baseline & 0.815 $\pm$ 0.018\\
\cline{2-3}
& $\tilde{R}_{\text{avg},t}$, fixed validation set & 0.890 $\pm$ 0.017\\
\cline{2-3}
& $\tilde{R}_{\text{w},t}$, weighted validation set & 0.893 $\pm$ 0.018\\
\cline{2-3}
& $\tilde{R}_{\text{sel},t}$, selective validation set & 0.865 $\pm$ 0.014\\
\hline
\end{tabular}
\caption{Comparison of results on the controller-selected holdout set.}
\label{tab:res}
\end{table}

\begin{figure}[!ht]
  \centering
  \subfloat[Prostate presence classification task]{\includegraphics[width=0.5\textwidth]{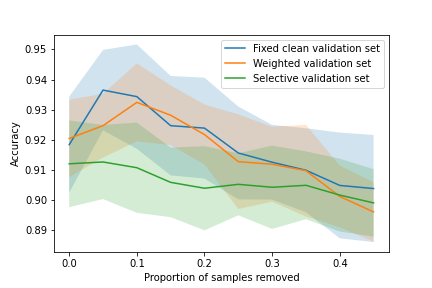}\label{fig:acc_reject}}
  \hfill
  \subfloat[Prostate segmentation task]{\includegraphics[width=0.5\textwidth]{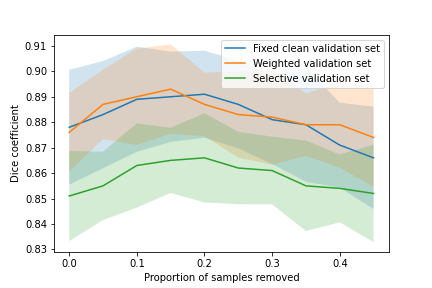}\label{fig:dice_reject}}
  \caption{Plots of the task performance (in respective Acc. and Dice metrics) against the ratios of removed holdout samples in each tasks.}
  \label{fig:perf_reject}
\end{figure}

To evaluate the task performances on the trained-controller-selected holdout set, the Acc. and Dice are summarised in Table \ref{tab:res}. The average training time was approximately 12 hours on a single Nvidia Quadro P5000 GPU. In both tasks, all three proposed RL-based IQA algorithms provide statistically significant improvements, compared with the non-selective baseline counterparts, with all \textit{p-values$<$0.001}. For both tasks, the results from the reward definition based on the selective validation set led to relatively inferior performances compared with the other two reward definitions, with statistical significance (\textit{p-values$<$0.001}). Interestingly, no statistical significance was found between the reward definitions based on fixed- and weighted validation sets, for the classification (\textit{p-value=0.06}) or segmentation (\textit{p-value=0.49}) tasks, despite the disagreement summarised in Fig. \ref{fig:cm}. Fig. \ref{fig:acc_reject} and \ref{fig:dice_reject} plot mean performance against (holdout) rejection ratio for the three reward computation strategies. The peak classification Acc. are $0.935$, $0.932$ and $0.913$ at $5\%$, $10\%$ and $5\%$ rejection ratios, for the fixed-, weighted- and selective reward formulations, respectively, while the peak segmentation Dice are $0.891$, $0.893$ and $0.866$ at $20\%$, $15\%$ and $20\%$ rejection ratios, respectively.

\section{Discussion and Conclusion}

An interesting observation when inspecting Fig. \ref{fig:perf_reject} is that, in both tasks, the task performance peaked before decreasing as more samples were discarded for most tested methods. This seems counter-intuitive as the controller was trained to select task amenable data. While it remains an open question, we consider the following potential contributing factors: the variance of predictions, the possible over-fitting of the RL algorithms, the potentially non-monotonic relation between the optimal predictions conditioned on different values of $s^{rej}$, and the limitation of the datasets which may be considered of above-average quality (therefore higher amenability that limits potential performance improvement). Importantly, the significant improvement over the non-selective baseline networks demonstrated the efficacy of the proposed IQA approach. 

The proposed weighted and selective reward formulations learned effective IQA without human labels of task amenability, which can be subjective and costly. Although the selective strategy performed moderately in this experiment, it may not be a general case for different datasets or applications and potentially provides a means to specify the desirable rejection rate.

In summary, this paper has formulated IQA as a measure of task amenability, which can be learned by the proposed RL algorithm with and without human labels. The proposed IQA has been demonstrated and analysed with experiments based on clinical ultrasound images from prostate cancer patients.

\section*{Acknowledgements}

This work is supported by the Wellcome/EPSRC Centre for Interventional and Surgical Sciences [203145Z/16/Z], the CRUK International Alliance for Cancer Early Detection (ACED) [C28070/A30912; C73666/A31378], EPSRC CDT in i4health [EP/S021930/1], the Departments of Radiology and Urology, Stanford University, the Natural Sciences and Engineering Research Council of Canada Postgraduate Scholarships-Doctoral Program (ZMCB), the University College London Overseas and Graduate Research Scholarships (ZMCB), GE Blue Sky Award (MR), and the generous philanthropic support of our patients (GAS).

\printbibliography

\end{document}